\documentclass[sigconf]{acmart}
\usepackage{booktabs}
\usepackage{graphicx}
\usepackage[normalem]{ulem}
\pdfoutput=1
\useunder{\uline}{\ul}{}

\renewcommand\footnotetextcopyrightpermission[1]{} 
\begin{document}

\title[Aggregate-Eliminate-Predict]{Aggregate-Eliminate-Predict: Detecting Adverse Drug Events from Heterogeneous Electronic Health Records}

\author{Maria Bampa}
\affiliation{%
  \institution{Dept. of Computer and Systems Sciences}
  \city{Stockholm}
  \country{Sweden}
}
\email{maria.bampa@dsv.su.se}

\author{Panagiotis Papapetrou}
\affiliation{%
  \institution{Dept. of Computer and Systems Sciences}
  \city{Stockholm}
  \country{Sweden}
}
\email{panagiotis@dsv.su.se}

\renewcommand{\shortauthors}{Bampa and Papapetrou, et al.}

\begin{abstract}
We study the problem of detecting adverse drug events in electronic healthcare records. The challenge in this work is to aggregate heterogeneous data types involving diagnosis codes, drug codes, as well as lab measurements. An earlier framework proposed for the same problem demonstrated promising predictive performance for the random forest classifier by using only lab measurements as data features. We extend this framework, by additionally including diagnosis and drug prescription codes, concurrently. In addition, we employ a recursive feature selection mechanism on top, that extracts the top-k most important features. Our experimental evaluation on five medical datasets of adverse drug events and six different classifiers, suggests that the integration of these additional features provides substantial and statistically significant improvements in terms of AUC, while employing medically relevant features.
\end{abstract}


\maketitle

\section{Introduction}
Adverse drug events (ADEs) refer to injuries caused by medication errors, allergic reactions or overdoses, and are related to drugs \cite{ADEs}. Before a medicine is released to the market, in clinical trials, a rigorous approach is followed to test their efficacy and safety in a rather limited cohort of patients \cite{pharmacovigilancefuture}. After the authorization and while the medicine is used from a large number of patients for an extended period of time, it can result to unwanted ADEs. More than half of the ADEs that lead to Emergency Department visits are preventable, but half of those are not reported since they are not easily identifiable \cite{burden}; hence resulting in unnecessary human suffering and a burden in the healthcare sector. An alternative approach is to employ machine learning methods in post-market surveillance, by exploiting the abundant information available in Electronic Health Records (EHRs) to identify ADEs that did not appear during the stage of clinical trials. 

More concretely, a patient case in an EHR consists of a number of medical variables, such as diagnoses (in the form of ICD10~\footnote{https://www.icd10data.com/ICD10CM/Codes} codes) and medications (in the form of ATC codes~\footnote{https://www.whocc.no/atc\_ddd\_index/}), lab measurements, and other clinical procedures that characterize the history of each patient. These could potentially act as highly informative features towards ADE prediction, and as as such, should be carefully taken into consideration in the learning process in order to make medically sound predictions. Nevertheless, EHR data are inherently sparse and heterogeneous, and as a result may contain a large fraction of empty values (see, e.g., Bagattini et al. \cite{theframework}), hence introducing critical challenges that need to be addressed in order to achieve stronger predictive models. 

Previous research in ADE prediction has focused on utilizing structured data or clinical text (e.g., \cite{Ztemporalweighting,Zpredictivemodelling,nlp1}) from EHRs and on extracting static features for learning predictive models. 

Zhao et al. utilize the structure of clinical codes (medication and diagnoses codes) and explores different ways to represent them \cite{Zconcepthierarchies, Zcascading}, while other related research consider the temporal aspect of clinical codes and measurements, and employ predictive modeling to identify previously unseen ADEs \cite{Zlearningtemporalweights, temporalityofclinicalevents, Ztemporalweighting}. Moreover, the use of natural language processing has been investigated for obtaining models that are able to predict unseen ADEs based on unstructured data, such as clinical text \cite{nlp1, nlp2}. Recent work by Bagattini et al. \cite{theframework} focuses on optimizing the random forest classifier for ADE prediction by taking into account the temporal nature and sparsity that characterizes clinical lab measurements, and employs them as the only predictors for ADE classification \cite{theframework}. 

To the best of our knowledge existing approaches for ADE prediction using EHRs have been mainly focusing on optimizing specific predictive models for a particular data source or data type. On the other hand, very limited attention has been given to building models that combine multiple heterogeneous data sources, while addressing feature sparsity and heterogeneity, as well as taking into consideration the nature of each predictor.

The main goal of this paper is, hence, to incorporate different heterogeneous EHR features that can improve ADE discovery. Specifically, this paper extends the framework proposed by Bagattini et al. \cite{theframework} by (1) including additional features and data types from a patient record, except for using only lab measurements, and (2) by benchmarking the proposed approach on six classification models instead of only the random forest classifier.  As shown in the experimental evaluation these additional features, i.e., drug and diagnoses codes, in conjunction with the clinical measurements as well as a recursive feature importance strategy, can have a detrimental effect in ADE detection, resulting in significant improvements in terms of AUC for several ADE types.

The \textbf{contributions} of this paper include: (a) the presentation of a workflow for ADE prediction using multiple data sources extracted from EHRs, that consists of two phases: feature aggregation, and recursive feature selection and predictive modeling; (b) the investigation of six classification models and eight feature integration approaches for ADE prediction; (c) an extensive experimental evaluation on the five most popular ADEs in the Stockholm EPR Corpus extracted from HealthBank \cite{healthbank} exploring all combinations of the three proposed feature types, as well as using six different classification models.

\section{Aggregate - Eliminate - Predict}
\label{sec:methods}

We first present the feature aggregation phase of our workflow, followed by the recursive feature elimination and predictive modeling phase in more detail.

\subsection{Feature Aggregation}
Let $\mathcal{E} = \{E_1, \ldots, E_n\}$ denote an EHR of a single patient, where each $E_i$ represents a medical event, and is represented by a triplet $(e, v, t)$, with $E_i.e \in \mathcal{T}$ being the event type over a set of possible event types $\mathcal{T} =\{t_1, \ldots, t_k\}$, $E_i.v$ being the event value, and $E_i.t$ the event timestamp.  In this paper, we consider two types of events, i.e., $k=2$: \emph{categorical} and \emph{continuous}.  Medical diagnoses and drug prescriptions (i.e., ATC and ICD10 codes) belong to the first case, while lab measurements belong to the second case.

Consider a set of EHRs denoted as $\mathcal{D} = \{\mathcal{E}^1,\ldots, \mathcal{E}^m\}$ over $m$ patients, and a time window $w=\{t_s,t_e\}$ defined at a particular location of interest in each $\mathcal{E}^i$, ending at time point $t_e$ (e.g., at the occurrence of an ADE) and spanning $w$ time points before $t_e$. The task is to extract a set of aggregated features over time for each $\mathcal{E}^i$.  Let $\mathcal{E}^i_{w} = \{E^i_{j}, \ldots, E^i_{j'}\}$ be the subsequence of events in $\mathcal{E}^i$ starting at timestamp $t_s$ and ending at $t_e$. Practically, $E^i_{j'}$ corresponds to the last event in $\mathcal{E}^i$ such that $E^i_{j'}.t \leq t_e$ and $E^i_{j}$ is the ``earliest'' event event in the time window $w$, such that $E^i_{j}.t \geq t_s$. 

For each distinct \emph{categorical} event $E_j$ in $\mathcal{E}^i_{w}$, we count how many times $E_j$ occurs in $w$, and assign this value as the feature value for this event. For the case of a \emph{continuous} event $E_j$, we apply the \textit{lr} transformation described in Baggatini et al. \cite{theframework}, such that $lr(E_j)$ maps the continuous feature to a real value capturing the underlying temporal structure in $E_j$. These transformations result into a tabular transformation of each $\mathcal{E}_i \in \mathcal{D}$, denoted as $\tau_w(\mathcal{E}_i)$, for a given window size $w$. 

This results in a new transformed EHR dataset containing aggregated features, using the above representation. This new aggregated dataset is denoted as $\hat{\mathcal{D}_w}= \{\tau_w(\mathcal{E}_1), \ldots, \tau_w(\mathcal{E}_m)\}$.

\subsection{Recursive Feature Elimination and predictive modeling}
The converted EHR dataset $\hat{\mathcal{D}_w}$ is passed to a \emph{recursive feature elimination} mechanism that recursively identifies the top-k most informative features.

Let $\mathcal{M}^{(0)} = f(\hat{\mathcal{D}_w})$ denote the model built using the aggregated dataset $\hat{\mathcal{D}_w}$ and a given predictive modeling function $f(\cdot)$. For example, $f(\cdot)$ can correspond to a decision tree induction algorithm or to the random forest algorithm. The main steps of this procedure are as follows: a significant level $\alpha$ is set, and at each iteration $i$:
\begin{itemize}
    \item the feature with the highest gini importance is identified, and if the value is greater than the significance level $\alpha$, the feature is removed;
    \item a new model, $\mathcal{M}^{(i)}$, built using the remaining feature set.
\end{itemize}
The process is repeated until the removal of any feature that reduces the AUC of the model in a validation set below a given threshold $\beta$ or $k$ features are selected. This results in the final model, which we denote as $\mathcal{M}^{(k)}$.

\begin{figure}[!t]
    \centering
    \includegraphics[width=0.9\columnwidth]{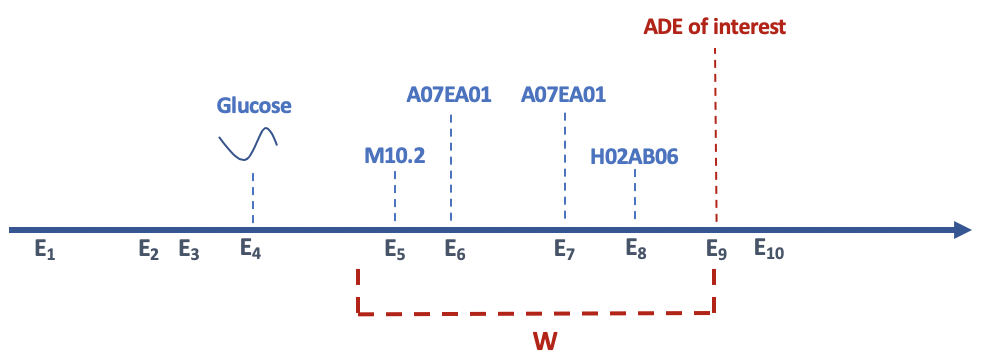}
    \caption{An EHR sequence and a window of interest $w$.}
    \label{fig:example}
\end{figure}

\section{Experimental Evaluation}
\label{sec:experiments}

\begin{table*}[!ht]
\caption{AUC obtained by 6 classifiers on 5 ADE datasets; The AUC reported is averaged over stratified 10-fold cross validation; in bold: the best AUC for each classifier; underlined: highest AUC for each ADE; In 8 integration approaches denoted as L:lab measurements, M:ATC, D:ICD-10, LM: labs and ATC, LD: labs and ICD-10, MD:ATC and ICD-10, LMD:labs, ATC and ICD-10.}
\centering
\scalebox{0.6}{
\resizebox{\textwidth}{!}{%
\begin{tabular}{@{}llllllllll@{}}
\toprule

\textbf{ADE} & \textbf{classifier} & \textbf{L} & \textbf{M} & \textbf{D} & \textbf{LM} & \textbf{LD} & \textbf{MD} & \textbf{LMD}& \textbf{LMD k best} \\ \midrule
\textbf{D611} & \textbf{RF100} & 0.8583 & 0.9226 & 0.8871 & 0.8965 & 0.8867 & \textbf{0.9311} & 0.9172 & 0.9166\\
\textbf{} & \textbf{SVMLin} & 0.8360 & 0.8338 & 0.7782 & 0.8429 & 0.8388 & 0.7804 & 0.8230 & \textbf{0.9137}\\
\textbf{} & \textbf{SVMPoly} & 0.8278 & 0.6314 & 0.4774 & 0.6642 & 0.6229 & 0.4451 & 0.4606 & \textbf{0.8897}\\
\textbf{} & \textbf{SVMrbf} & 0.8675 & 0.8073 & 0.7381 & 0.8568 & 0.8221 & 0.8028 & 0.8547 &\textbf{0.9137} \\
\textbf{} & \textbf{MLP} & 0.8554 & 0.8230 & 0.7566 & 0.8707 & 0.8170 & 0.8140 & 0.8573 & \textbf{0.9076}\\
\textbf{} & \textbf{XGB} & 0.8494 & 0.9149 & 0.8939 & 0.9184 & 0.8961 & 0.9306 & {\ul \textbf{0.9332}}  & 0.9243\\ \midrule
\textbf{G620} & \textbf{RF100} & 0.8376 & 0.7200 & 0.8166 & 0.8677 & 0.8625 & 0.8598 & 0.8890 &  \textbf{0.8982} \\
\textbf{} & \textbf{SVMLin} & 0.7037 & 0.6379 & 0.7314 & 0.7148 & 0.7476 & 0.7198 & 0.7746 &\textbf{0.8299}\\
\textbf{} & \textbf{SVMPoly} & 0.7198 & 0.6741 & 0.2787 & 0.7196 & 0.3674 & 0.3373 & 0.3898 &\textbf{0.7914}\\
\textbf{} & \textbf{SVMrbf} & 0.8070 & 0.7272 & 0.7619 & 0.8006 & 0.7667 & 0.7683 & 0.7967 &{\ul \textbf{0.9147}}\\
\textbf{} & \textbf{MLP} & 0.6577 & 0.7276 & 0.7505 & 0.7622 & 0.7667 & 0.7744 & 0.7800 &\textbf{0.8211}\\
 & \textbf{XGB} & 0.8351 & 0.7620 & 0.8242 & 0.8510 & 0.8595 & 0.8704 & 0.8770 & \textbf{0.8886}\\ \midrule
\textbf{T784} & \textbf{RF100} & 0.5939 & 0.7081 & 0.6032 & 0.7250 & 0.6591 & 0.7307 & {\ul \textbf{0.7616}} &0.7608\\
\textbf{} & \textbf{SVMLin} & \textbf{0.6443} & 0.6188 & 0.5996 & 0.6287 & 0.6080 & 0.6362 & 0.5974 &0.6417\\
\textbf{} & \textbf{SVMPoly} & 0.4685 & 0.4840 & 0.4608 & 0.4437 & 0.4601 & 0.4407 & 0.4161 &\textbf{0.5513}\\
\textbf{} & \textbf{SVMrbf} & 0.6377 & 0.7033 & 0.6079 & 0.7208 & 0.6167 & 0.6695 & 0.6877 &\textbf{0.7365}\\
\textbf{} & \textbf{MLP} & 0.5871 & 0.6358 & 0.5993 & 0.6237 & 0.5921 & 0.6340 & 0.6374 &\textbf{0.6864}\\
\textbf{} & \textbf{XGB} & 0.6298 & 0.7108 & 0.6146 & 0.7235 & 0.6316 & 0.7374 & 0.7417 &\textbf{0.7460}\\ \midrule
\textbf{T808} & \textbf{RF100} & 0.8910 & 0.8888 & 0.8679 & 0.9109 & 0.9293 & 0.9320 & 0.9320 & \textbf{0.9364}\\
\textbf{} & \textbf{SVMLin} & 0.7873 & 0.7058 & 0.7768 & 0.7967 & 0.8176 & 0.8315 & 0.8509 & \textbf{0.8721}\\
\textbf{} & \textbf{SVMPoly} & 0.8074 & 0.6930 & 0.7741 & 0.7729 & 0.7949 & 0.7861 & 0.8109 & \textbf{0.8927}\\
 & \textbf{SVMrbf} & 0.8903 & 0.8168 & 0.8112 & 0.8918 & 0.8453 & 0.8262 & 0.8547& {\ul \textbf{0.9389}}\\
 & \textbf{MLP} & 0.7841 & 0.7589 & 0.7867 & 0.8405 & 0.8432 & 0.8454 & 0.8825 & \textbf{0.9132}\\
 & \textbf{XGB} & 0.8900 & 0.8875 & 0.8894 & 0.9165 & 0.9078 & \textbf{0.9226} & 0.9204 & 0.9206\\ \midrule
\textbf{T887} & \textbf{RF100} & 0.6533 & 0.7308 & 0.7029 & 0.7832 & 0.7399 & 0.7652 & \textbf{0.7873} & 0.7868\\
 & \textbf{SVMLin} & 0.6797 & 0.6417 & 0.6275 & 0.6560 & 0.6540 & 0.6573 & 0.6744 & \textbf{0.7269}\\
 & \textbf{SVMPoly} & 0.6835 & 0.6457 & 0.4529 & 0.6169 & 0.4762 & 0.4873 & 0.3876 & \textbf{0.7356} \\
 & \textbf{SVMrbf} & 0.7002 & 0.7222 & 0.6682 & 0.7648 & 0.7130 & 0.7262 & 0.7518 &\textbf{0.7683}\\
 & \textbf{MLP} & 0.6280 & 0.6612 & 0.6530 & \textbf{0.7135} & 0.6930 & 0.6843 & 0.7046 &0.6989\\
 & \textbf{XGB} & 0.6813 & 0.7352 & 0.6773 & 0.7762 & 0.7353 & 0.7759 & 0.7932 & {\ul \textbf{0.7951}}\\ \bottomrule
\end{tabular}}%
}
\label{tab:my-table}
\end{table*}

\subsection{Data and Setup}
The medical database consists of information about diagnoses, medications, blood and laboratory values for 1,314,646 patients obtained from Health Bank at Stockholm University~\cite{healthbank}; an anonymized patient record from the TakeCare CGM patient record system used at Karolinska University Hospital in Stockholm, Sweden. Structured data are labelled using common encoding systems such as the Anatomical Therapeutic Chemical Classification System (ATC) for medications, the International Statistical Classification of Diseases and Related Health Problems, 10th Edition (ICD-10) for diagnoses as well as the Nomenclature, Properties and Units (NPU) coding system~\footnote{http://www.ifcc.org/ifcc-scientific-division/sd-committees/c-npu/} for clinical laboratory measurements.

The performance of the proposed workflow has been evaluated using the following benchmarked classification algorithms: (1)Random Forests with 100 trees and gini impurity as the split criterion, Support Vector Machines(SVM) with weighted class balance and (2)linear kernel, (3)polynomial kernel, (4) RBF kernel, (5) a Multi-layer-percepton(MLP) with 3 hidden layers and (6)eXtreme Gradient Boosting(XGB) with 100 trees. All models were trained in the 8 feature integration approaches. Since the datasets in this study are imbalanced, the evaluation metric we used is the Area Under the ROC Curve (AUC), as it has been shown to be an appropriate metric for imbalanced datasets, not biased towards the majority class \cite{theframework,ROC} and was obtained from stratified 10-fold cross-validation.   

\subsection{Results}
Table 1 presents our experimental findings in terms of predictive modelling. The first experiment focused on extending the work of Bagattini et al. \cite{theframework} incorporating features of other types while only using Random Forests, for easier comparison with the proposed framework. It can be seen that for the the integrated set of lab measurements, medication and diagnoses codes (LDM), Random Forest consistently outperformed the integration approach with only the lab values (L). Specifically, 4 out of 5 studied ADEs showed an improvement, in terms of mean AUC over stratified 10-fold cross-validation, ranging from 4 to 10 \%. This indicates that a pool of different medical features can constitute promising ADE predictors, even though the dimensionality of the feature space increases. 

Furthermore, we provide a statistical analysis for the 7 integration approaches of the 5 ADEs, using the Random Forest classifier, to identify if the investigated approaches perform equally well. We use the Friedman test followed by Nemenyi post-hoc test when the null hypothesis is rejected \cite{statisticalcomparison}. The Friedman test returned a p value of 2.922e-05 indicating that among the proposed integration approaches there is at least one that statistically differs from the others. Figure 2 depicts the results of the post-hoc Nemenyi test where it can be observed that the LMD integration approach is statistically different from the datasets that contain the lab measurements(L), the diagnoses codes(D) as well as the fusion of lab and diagnoses codes(LD). 

Next, we investigate the effect of other classifiers in all the integration approaches. Before feature elimination we observe that in the LMD integration approach, for ADEs D61.1 and  T88.7, XGB yielded slightly better results, with random forest being the winner in G62.0, T78.4 and T80.8. The last column of table 1 presents the mean AUC obtained for the k best features after feature elimination. For D61.1 and G62.0 k=55, for T78.4 k=270, for T80.8 k=190 and lastly for T88.7 k=120, where k is the number of optimal features. It can be observed that across most classifiers and for all studied ADEs the feature elimination approach notably yielded better results. Note that when comparing the last two columns(LMD and LMD k best) it can be seen that the SVM classifiers had a noteworthy increase in mean AUC, while Random Forest and XGB in most cases slightly improved.

\begin{figure}[!ht]
    \centering
 \centerline{\includegraphics[width=\columnwidth]{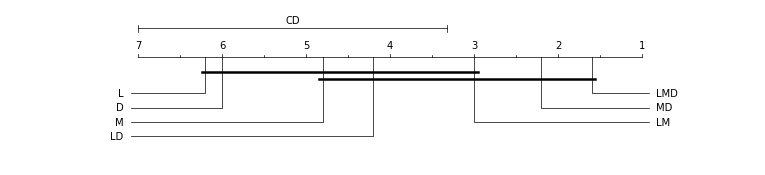}}
 \caption{post-hoc Nemenyi test for multiple comparison of the 7 integration approaches using the Random Forest classifier, excluding the one obtained by RFE.}
    \label{fig:my_label}
\end{figure}

\subsection{Medical relevance}
Feature importance for the integrated datasets that contain the lab measurements, medication and diagnoses codes, before and after feature elimination, indicated the following among the most important features chosen by the Random Forest classifier: NPU03568,  NPU01944, A04AA01, Z51.1. According  to  the  literature,  irregular levels of NPU03568 (Thrombocytes/platelets) and NPU01944 (Erythrocytes) ~\footnote{ https://www.cancer.gov/publications/dictionaries/cancer-terms/def/erythrocyte} are indicators of aplastic anaemia ~\footnote{https://www.niddk.nih.gov/health-information/blood-diseases/aplastic-anemia-myelodysplastic-syndromes}, Z51.1 (Encounter  for  antineoplastic  chemotherapy) is a cancer treatment related to/can cause aplastic anaemia  ~\footnote{https://medlineplus.gov/aplasticanemia.html} and lastly drug code A04AA01 (ondansetron, in the category of antiemetics and antinauseants) was found to be a drug administered to patients that undergo chemotherapy to prevent nausea and vomiting ~\footnote{https://www.fda.gov/drugs/postmarket-drug-safety-information-patients-and-providers/ondansetron-marketed-zofran-information}.

\begin{figure}[!ht]
    \centering
 \centerline{\includegraphics[width=0.8\columnwidth]{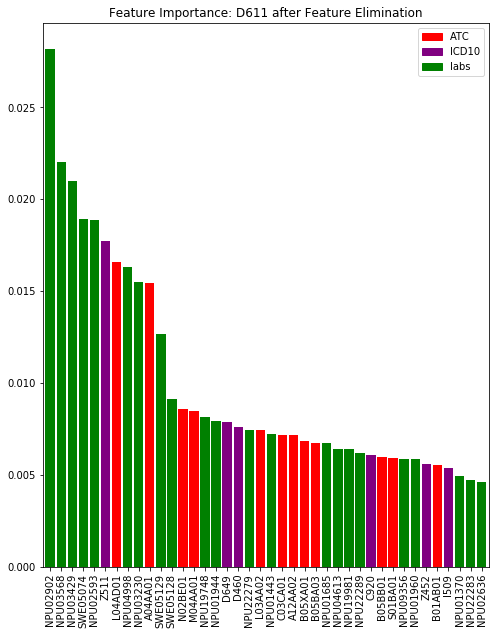}}
 \caption{Feature importance on D61.1 after feature elimination}
    \label{fig:my_label}
\end{figure}

\section{Conclusions}
\label{sec:conclusions}

We presented a workflow for predicting ADEs from EHRs using multiple disparate and complex medical data sources. We demonstrated the importance of incorporating different heterogeneous EHR features in the learning process, compared to only using lab measurements  as it can improve ADE discovery and lead to medical sound predictions. Furthermore, we illustrated the importance of feature elimination and employing medically relevant features as it can substantially improve the classification task. Future research should focus on incorporating other types of features such as clinical text, investigate the temporal aspect of all features used, as well as study the effect of dynamically choosing different patient history lengths per feature. 

\begin{acks}
This work was partly supported by the VR-2016-03372 Swedish Research Council Starting Grant. Ethical approval was granted by the Stockholm Regional Ethical Review Board under permission no. 2012/834-31/5.
\end{acks}

\bibliographystyle{ACM-Reference-Format}

\end{document}